\documentclass[10pt,twocolumn,letterpaper]{article}

\usepackage{cvpr}
\usepackage{times}
\usepackage{epsfig}
\usepackage{graphicx}
\usepackage{amsmath, nccmath}
\usepackage{amssymb}
\usepackage{amsthm}

\usepackage[linesnumbered, ruled,vlined]{algorithm2e}
\usepackage{algpseudocode}

\let\svthefootnote\svthefootnote

\usepackage[pagebackref=true,breaklinks=true,letterpaper=true,colorlinks,bookmarks=false]{hyperref}

\cvprfinalcopy 

\def\cvprPaperID{7392} 

\ifcvprfinal\fi
\begin{document}

\title{Adjusting Decision Boundary for Class Imbalanced Learning}

\author{Byungju Kim \\
KAIST\\
{\tt\small byungju.kim@kaist.ac.kr}
\and
Junmo Kim*\\
KAIST\\
{\tt\small junmo.kim@kaist.ac.kr}
}

\maketitle

\begin{abstract}
Training of deep neural networks heavily depends on the data distribution.
In particular, the networks easily suffer from class imbalance.
The trained networks would recognize the frequent classes better than the infrequent classes.
To resolve this problem, existing approaches typically propose novel loss functions to obtain better feature embedding.
In this paper, we argue that drawing a better decision boundary is as important as learning better features.
Inspired by observations, we investigate how the class imbalance affects the decision boundary and deteriorates the performance.
We also investigate the feature distributional discrepancy between training and test time.
As a result, we propose a novel, yet simple method for class imbalanced learning.
Despite its simplicity, our method shows outstanding performance.
In particular, the experimental results show that we can significantly improve the network by scaling the weight vectors, even without additional training process.
\vspace{-3 mm}

\end{abstract}


\section{Introduction}
\let\thefootnote\relax\footnote{Code available: \href{https://github.com/feidfoe/AdjustBnd4Imbalance}{https://github.com/feidfoe/AdjustBnd4Imbalance}}
Data is imbalanced in nature.
We frequently encounter everyday information, while we rarely face singular information.
Despite this imbalance, humans do not have any trouble in learning and recognizing things.
Also, we often learn from rare, but intense experiences.
However, when it comes to the domain of machine learning, the imbalance of data becomes a critical issue.
It deteriorates the performance of the trained machines.
Especially in deep neural networks, imbalanced data distribution is highly critical since they learn directly from the data distribution.
To this end, many of the widely used public datasets provide well-balanced class distribution.
In other words, their collectors have discarded a large amount of data from frequent classes to adjust the class balance.
It is clearly wasteful and redundant in terms of both information and human efforts.

The optimal scenario is certainly to train a machine using every data that we have access to.
However, the disparity between samples often induces a disparity between the accuracy of classes.
Features are often biased toward frequently appearing classes, so the less frequent classes have poor feature representation.
Consequently, a trained machine recognizes frequent classes, whereas it shows poor performance with infrequent classes.
Understanding how such a phenomenon is developed in class imbalanced learning could provide a novel viewpoint to mitigate the problem of the imbalanced performance.
In this work, we provide an in-depth analysis based on observation and propose a simple but powerful method for class imbalanced learning.

One important observation is that minimizing the empirical loss with a conventional training framework results in decision boundaries that allocate a larger volume of the feature space to more frequent classes.
This suggests that the decision boundary is biased toward less frequent classes.
Furthermore, we show that the bias in the decision boundary is closely related to the norm of each weight vector.
Low sample frequency reduces the norm of the weight vector and leads to a disadvantageous decision boundary.
Therefore, we propose the Weight Vector Normalization (WVN) method to draw the decision boundary at the middle of the weight vectors.

Another motivational observation is regarding how the features of each class are distributed.
If a network is trained for recognition, the features of each class form a cluster in the feature space.
In the image space, the size of each cluster follows the sample frequency; more samples literally form a larger cluster.
However, we have observed that the more frequent classes rather form smaller clusters in feature space; more samples induce a higher density.
This \textit{size reversal} of clusters is due to the disparity of generalization.
A trained neural network is more generalized to frequent classes, whereas it is over-fitted to infrequent classes.
This suggests that a larger margin is required for less frequent classes.
To resolve this problem, we propose a weight re-scaling method (RS).
Once the network is done training, we adjust the decision boundary depending on the sample frequency.
Despite the simplicity of the proposed method, it shows outstanding performance.
Interestingly, we achieved better performance than the existing methods without using any additional training process.
This suggests that we can obtain a feature extractor of fine quality by minimizing empirical loss and that the problems with class imbalanced learning mainly lie in how to draw the appropriate decision boundary. 

Our main contributions can be summarized as follows:
Firstly, we present an in-depth analysis on class imbalanced learning, in terms of the norm of the weight vector.
Our analysis shows that there is obvious correlation between the norm and sample frequency.
Secondly, we show that we can adjust the decision boundary by controlling the norm of the weight vector.
With concordant observations, we propose a novel method, which outperforms the previous methods.
Lastly, we experimentally show that the features from our baseline network are already of fine quality; hence we can achieve better performance than the existing methods with a delicately drawn decision boundary.

\section{Related Works}
\label{sec:related_works}
The great majority of existing algorithms used to resolve the data imbalance problem can be categorized as either re-sampling or re-weighting.
The data re-sampling approach is intuitively straight forward and relatively simple: \textit{``Since we have imbalanced number of data for each class, duplicate or discard what we already have."}
Properly over-sampled \cite{smote, han2005borderline, bunkhumpornpat2009safe, barua2012mwmote, ramentol2012smote, he2009learning}, or under-sampled \cite{japkowicz2002class, haixiang2017learning, buda2018systematic, lee2016plankton} data makes a model perform better.
However, both over-sampling and under-sampling approach have notable weaknesses.
The over-sampling method causes a model to become over-fitted to the duplicated samples. 
To minimize the over-fitting problem, SMOTE~\cite{smote} and its variants~\cite{han2005borderline, bunkhumpornpat2009safe, barua2012mwmote, ramentol2012smote} have been proposed to generate samples of infrequent classes.
The recently proposed generative adversarial networks\cite{goodfellow2014generative,yin2017semi,douzas2018effective} can also resolve this problem.
However, it is difficult to overcome the fundamental deficiency in data samples.

On the other hand, the under-sampling approach easily deteriorates the overall performance, struggling with severer data deficiency.
As Sun \etal described in \cite{sun2017revisiting}, the performance of neural networks logarithmically increases based on the volume of training data.
This implies that discarding samples is critical in terms of overall performance.
In \cite{lee2016plankton}, the authors pointed out that the natural distribution is also a valuable information, so we need to fully exploit the data.
For this reason, the over-sampling strategy is preferred to under-sampling.

The re-weighting approach is also considered as a cost-sensitive approach.
The underlying concept of a cost-sensitive approach for class imbalanced learning is to treat different predictions differently.
In \cite{elkan2001foundations}, the authors researched weighting methods for the binary classification task.
Similarly, a cost-sensitive SVM for highly imbalanced datasets was proposed in \cite{tang2008svms, yan2017optimizing, lee2017instance}.
To obtain a better performing model, the ensemble method was adopted to both cost-sensitive \cite{zadrozny2003cost}, and sampling approaches \cite{wang2014resampling,wang2017svm}.

Following the explosive development of CNN-based models, deep learning based algorithms that resolve the class imbalance problem have been proposed.
Under excessive class imbalance, re-weighting the classification loss due to the inverse of the sample frequency can make a network diverge during training.
To this end, Cui \etal proposed the concept of \textit{effective number} of samples to re-balance the classification loss \cite{cui2019class}.
In advance of \cite{cui2019class}, Lin \etal proposed focal loss\cite{lin2017focal}, which weights the classification loss depending on the prediction results.
Focal loss helps the network to focus on poorly predicted samples and not become over-fitted to the well-predicted samples.
Cao \etal proposed label-distribution-aware margin loss\cite{cao2019learning} which aims to generalize the minority class better, by considering the label distribution in the loop.
Khan \etal proposed a novel loss function by estimating the uncertainty of each class\cite{khan2019striking}.
In \cite{huang2016learning, dong2018imbalanced}, the authors also proposed another form of loss to train the neural networks by sampling neighbors.

As enumerated above, deep learning based methods mainly focus on studying a novel loss function for class imbalanced learning.
Unlike these researches, our work involves neither re-sampling nor re-weighting.
Our proposed method regulates the neural network and adjusts the decision boundary based on the sample frequency of each class.
Similar to \cite{cao2019learning}, we analyze the class imbalance problem in terms of generalization.
We compare the generalization for each class and use the analysis as a prior information to adjust the decision boundary.
The details of the method are presented in the following section with a thorough justification.

\section{Method}
Before describing the method, we define the notations and overall framework.
If it is not specifically mentioned, the notations that appear in this paper refer to the following.

\subsection{Empirical Loss Minimization}
Suppose that we have ㅁ training dataset $\mathcal{D}=\{(x_{i}, y_{i})\}_{i=1}^{N}$ of $N$ image-label pairs, where the label space is $\{1, ..., K\}$; it is a classification problem with $K$ classes.
Since our target task is class imbalanced learning, we further segment the dataset as $\mathcal{D} = \bigcup_{j=1}^{K}\mathcal{D}_{j}$, where $\mathcal{D}_j$ is a subset of the whole dataset, which consists of samples from class $j$.
Then, we define $n_j$ as the number of samples in $\mathcal{D}_j$.
Without loss of generality, we can set $n_{1} \geq ... \geq n_{K}$.
Following prior research \cite{cui2019class}, we define the imbalance ratio of the dataset as $n_{1} / n_{K}$.

For training, we employ a general framework.
We first feed an input image $x$ into a feature extraction network $f(\cdot)$.
It outputs a feature vector, $f(x) \in \mathbb{R}^d$.
Then, a classifier, which consists of single fully connected layer, outputs a logit vector, $l(x) \in \mathbb{R}^K$, by calculating the inner-product between $f(x)$ and the learnable parameter, $W \in \mathbb{R}^{d\times K}$.
We can write $W$ in a vector form as $W = [w_{1}, ... ,w_{K}]$, where $w_{j} \in \mathbb{R}^d$ is a \textit{weight vector} for class $j$.
The operation of the classifier can be written as follows:
\begin{equation}
\begin{split}
l(x) & = W^{T}f(x) \\
     & = [w_{1}^{T}f(x); ... ;w_{K}^{T}f(x)].
\end{split}
\label{eq:lx}
\end{equation}
For brevity, we drop the additive bias term.
Note that we are considering a linear classifier.
Then, we apply softmax operation to convert $l(x)$ into a vector of probabilities, $p(x)$.
Each element of $p(x)$ represents the probability of input $x$ belonging the corresponding class.
We compute the cross-entropy loss between the one-hot encoded ground truth label and $p(x)$, so that we can calculate the gradients for the learnable parameters.

The described framework trains the neural network by minimizing the empirical loss.
Given a dataset, $\mathcal{D}$, the empirical loss can be formulated as follows:
\begin{equation}
\vspace{-2 mm}
\mathcal{L}(\mathcal{D}) = \frac{1}{|\mathcal{D}|}\sum_{x\in \mathcal{D}} \ell(y,x),
\label{eq:define_loss}
\end{equation}
where $|\mathcal{D}|$ denotes the size of the dataset, and $\ell(\cdot,\cdot)$ denotes the cross entropy loss between the label and $p(x)$.
Considering $\mathcal{D}$ to be a union of $\mathcal{D}_j$, we can rewrite the empirical loss as the weighted summation of the class-wise empirical loss as follows:
\begin{equation}
\vspace{-2 mm}
\mathcal{L}(\mathcal{D}) = \sum_{j=1}^{K}\frac{n_{j}}{N}\mathcal{L}(\mathcal{D}_{j}). 
\label{eq:sum_loss_j}
\end{equation}
From Eq.(\ref{eq:sum_loss_j}), it can be seen that minimizing $\mathcal{L}(\mathcal{D})$ is highly likely to result in $\mathcal{L}(\mathcal{D}_{1}) \leq \mathcal{L}(\mathcal{D}_{2}) \leq ... \leq \mathcal{L}(\mathcal{D}_K)$, if the number of samples for each class is highly imbalanced.
The asymmetrically optimized class-wise empirical loss is likely to result in a decision boundary that is biased toward less frequent classes~\cite{khan2019striking}.
We consent to the analysis of \cite{khan2019striking}; however, we focus more on the norm of each weight vector, unlike the authors who focused on the directions.

\begin{figure}[t]
  \centering
  \includegraphics[width=\linewidth]{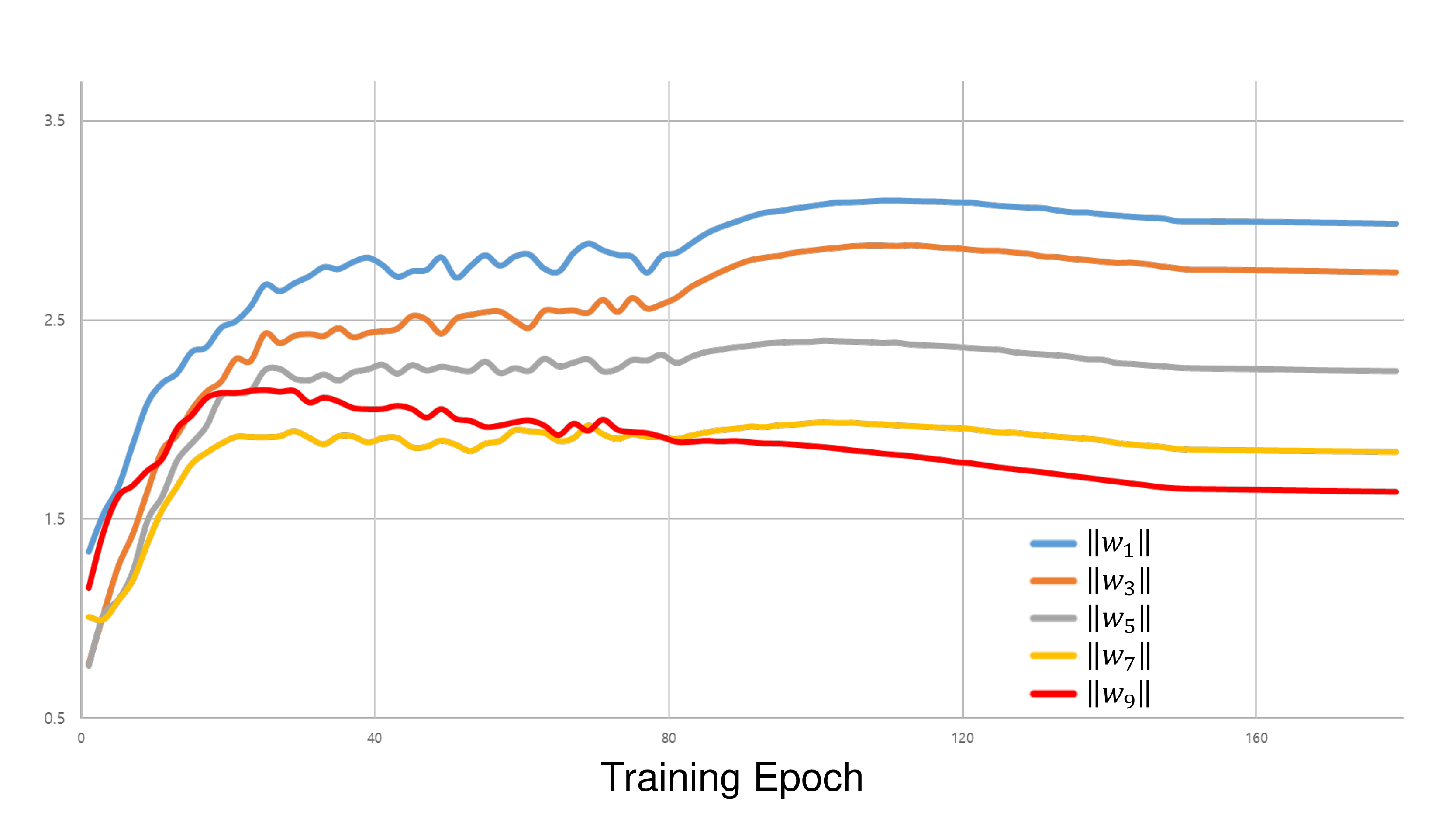}
  \caption{How the norm changes during the training process. 
  Note that class 1 is the most frequent class, while class 9 is the least frequent class in the figure.
  Early in training, the norms do not show clear correlation with the sample frequency. 
  However, in the later stage, the norm of each class is aligned with the sample frequency}
\vspace{-2 mm}
  \label{fig:norm_vs_epoch}
\end{figure}

\begin{figure}[t]
  \centering
  \includegraphics[width=\linewidth]{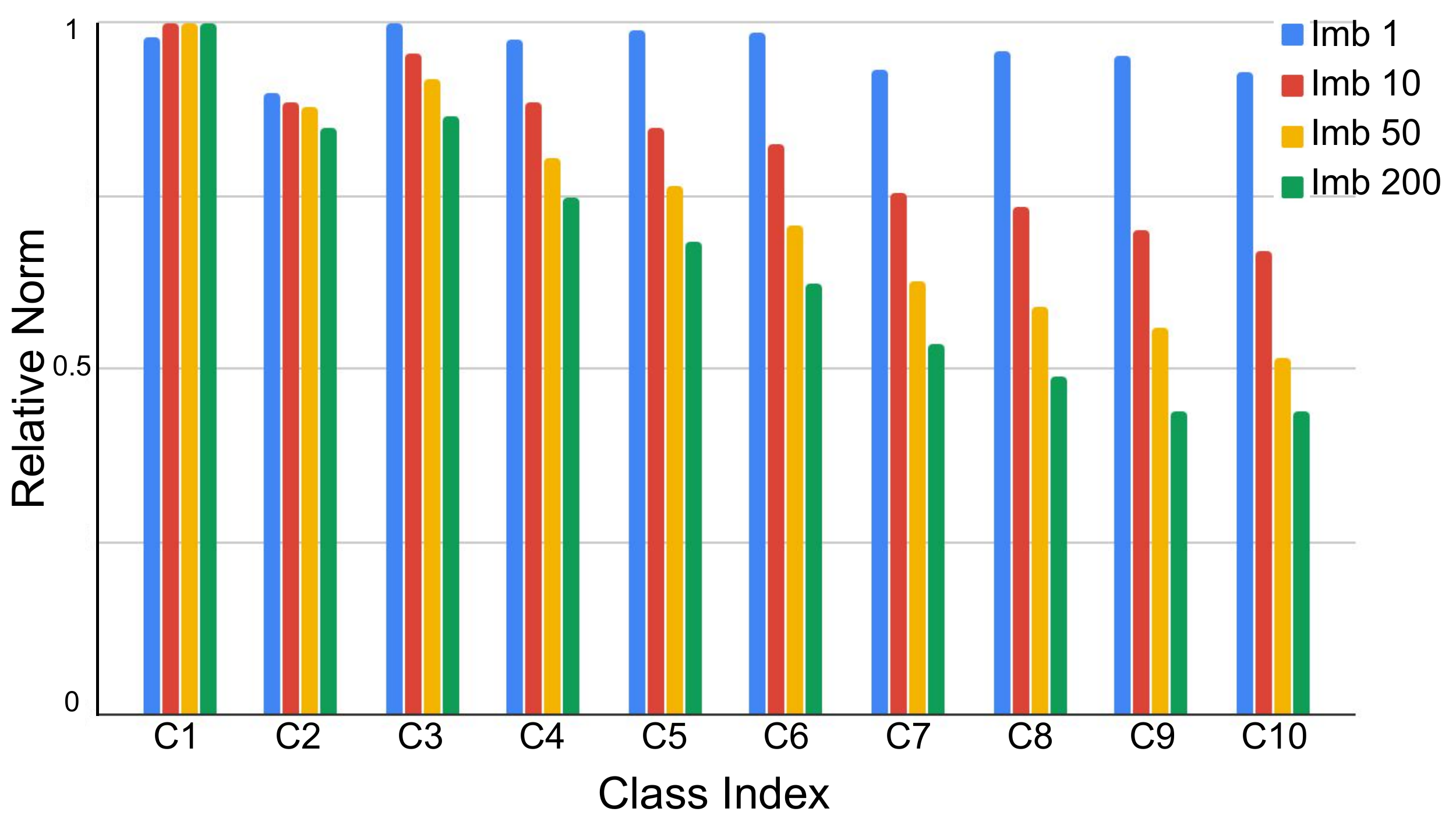}
  \caption{Relative norm of the weight vectors depending on the imbalance ratio. 
  Except for the case of Imb1, the weight vector for the most frequent class has the largest norm, whereas the weight vector for the least frequent class has the smallest norm.
  Moreover, if the data is more imbalanced, larger discrepancy appears in the norm}
\vspace{-2 mm}
  \label{fig:num_sample_vs_norm}
\end{figure}

\subsection{Norm and Decision Boundary}
\label{sec:imb_and_norm}
We start with an observation on the tendency of the norm of each weight vector.
Figure~\ref{fig:norm_vs_epoch} shows how the norm of each weight vector changes over the training process.
Early in the training, the norms do not show a clear correlation with the sample frequency, since it is suffering even for the training data.
Notably, the norms of every weight vector are increasing.
During the later stage of the training, the graph of the norms become disentangled, presenting an apparent correlation.
Figure~\ref{fig:num_sample_vs_norm} illustrates the relative norm of the weight vectors.
The norms are relatively uniform, if the training data is well-balanced (Imb 1).
Since $||w_k||_2$ can be interpreted as a multiplicative bias, the fluctuation presents a natural variation of the bias.
However, when the training data contains an imbalance, $||w_1||_2$ has the largest value while $||w_{K}||_2$ has the smallest value.
Figure~\ref{fig:num_sample_vs_norm} presents the evident correlation between the norm and the sample frequency.
With a more imbalanced sample frequency, the weight vectors of the classifier are more imbalanced in terms of the norm.
This observation suggests that a high sample frequency causes a large norm of the weight vector.

We can understand this tendency by investigating the partial derivative of $\mathcal{L}(\mathcal{D}_{j})$ with respect to $||w_k||_2$.
Consider a sample, $x \in \mathcal{D}_{j}$.
Since the $k$-th element of $l(x)$ can also be expressed as $w^{T}_{k}f(x) = ||w_{k}||_{2}\: ||f(x)||_{2}\cos(\theta)$, the partial derivative can be formulated as follows:
\begin{equation}
\begin{split}
\frac{\partial \ell(j,x)}{\partial ||w_k||_{2}} &= 
\frac{\partial \ell(j,x)}{\partial l(x)}\ \frac{\partial l(x)}{\partial ||w_k||_{2}} \\ 
= & \left\{  \begin{array}{l} 
p^{k}(x)||f(x)||_{2}\cos(\theta_{k}^{x}) \quad\quad\ \ \ \text{if }k\neq j\\
(p^{k}(x)-1)||f(x)||_{2}\cos(\theta_{k}^{x}) \  \text{if }k= j\\
  \end{array}
  \right.
,
\end{split}
\label{eq:partial_derivative}
\end{equation}
where $p^{k}(x)$ denotes the $k$-th element of $p(x)$, and $\theta_{k}^{x}$ denotes the angle between $f(x)$ and $w_k$.
Eq.(\ref{eq:partial_derivative}) shows that the sign of $\partial \ell(j,x)/\partial ||w_k||_{2}$ is dependent on $\theta_{k}^{x}$, since the other terms always have a fixed sign.
Once the network is sufficiently trained, so that the empirical loss has been sufficiently minimized, $\cos(\theta_{k}^{x})$ is highly likely to have a positive value if $k = j$ for all $x \in \mathcal{D}_j$.
This suggests that $\partial  \mathcal{L}(\mathcal{D}_{j})/\partial ||w_j||_2$ has a negative value, so $||w_j||_2$ should be increased by minimizing $\mathcal{L}(\mathcal{D}_{j})$.
On the other hand, if $k \neq j$, $\partial \mathcal{L}(\mathcal{D}_{j})/\partial ||w_k||_2$ can be either positive or negative depending on the correlation between classes $j$ and $k$.
Assuming a highly imbalanced sample frequency, Eq.(\ref{eq:sum_loss_j}) and Eq.(\ref{eq:partial_derivative}) imply that $||w_{1}||$ is likely to have the largest value among the weight vectors.

\begin{figure}[t]
  \centering
  \includegraphics[width=\linewidth]{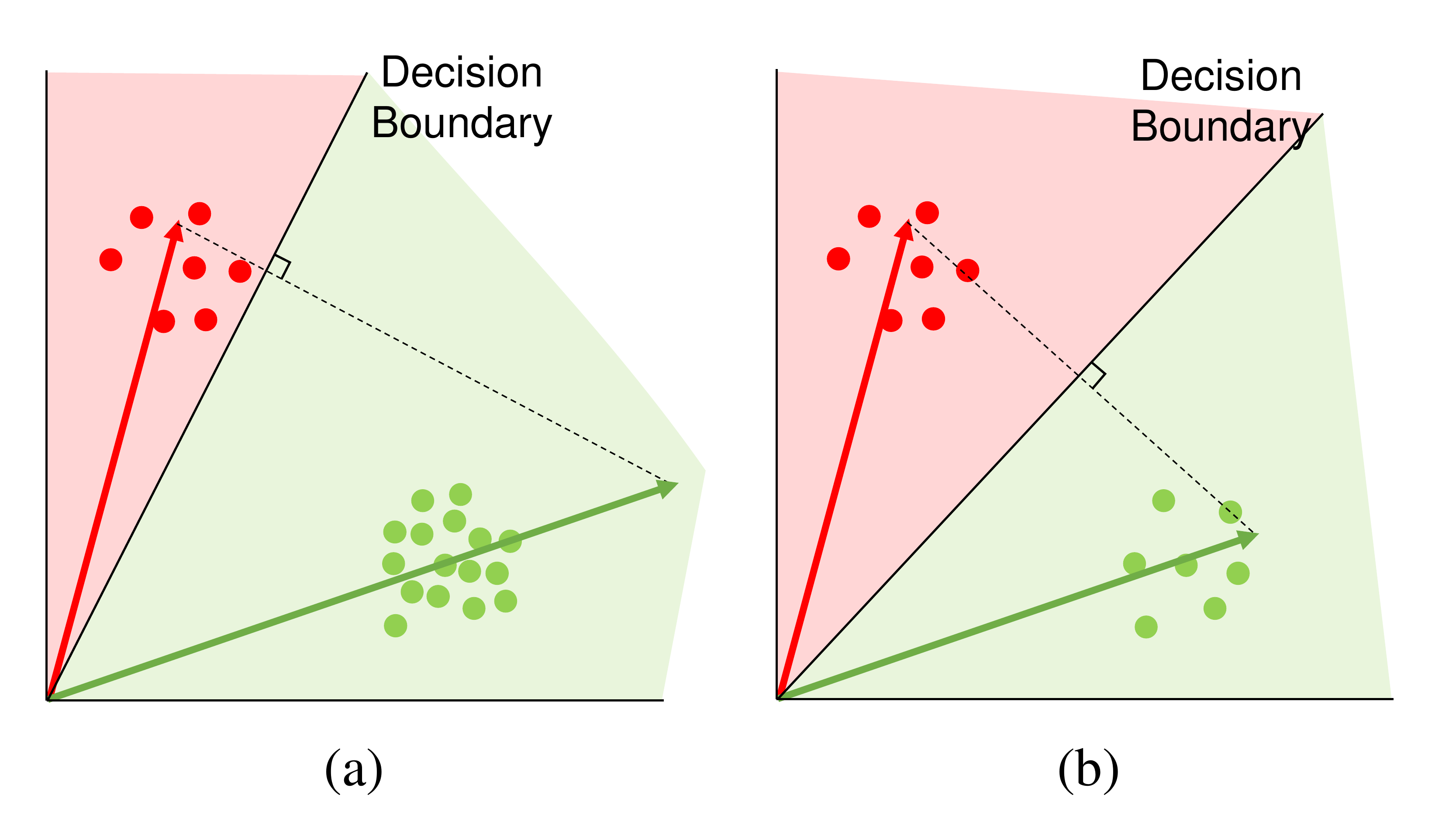}
\vspace{-3 mm}
  \caption{Correlation between the decision boundary and the weight vectors.
  (a) If two weight vectors have different norms, the decision boundary is drawn leaning toward the weight vector with the smaller norm.
  (b) If they have identical norms, the decision boundary is drawn at the middle.
  This figure also shows that we can adjust the decision boundary by adjusting the norm
  }
  \label{fig:norm_and_boundary}
\vspace{-3 mm}
\end{figure}

The norm of the weight vector and the decision boundary are closely related.
In the feature space, the decision boundary between class $i$ and $j$, is a set of points that satisfy $w_{i}^{T}f(x) = w_{j}^{T}f(x)$;
we can rewrite this hyperplane as follows:
\begin{fleqn}
\begin{equation}
B(i,j) = \{x\in \mathbb{R}^{d}| ||w_{i}||_{2}\cos(\theta_{i}^{x}) = ||w_{j}||_{2}\cos(\theta_{j}^{x}) \}.
\label{eq:decision_boundary}
\end{equation}
\end{fleqn}
This implies that the weight vector of larger norm would form wider angle with the decision boundary.

Figure~\ref{fig:norm_and_boundary} illustrates how the norm of each weight vector affects the decision boundary.
Although the direction of each weight vector is fixed, the boundary changes depending on the norm.
If we train the network without any regularization, the weight vectors are formed as shown in Figure~\ref{fig:norm_and_boundary} (a);
the weight vector for more frequent class has a larger norm, so the decision boundary is biased toward the less frequent class.
As a result, a smaller volume of the feature space is allocated to the less frequent class.
On the other hand, Figure~\ref{fig:norm_and_boundary} (b) shows that the decision boundary is drawn at the middle of the two weight vectors since they have similar norms.
As illustrated in Figure~\ref{fig:num_sample_vs_norm}, well-balanced sample frequency brings about well-balanced norm.
As a result, comparable volumes of feature space are allocated to each class.
To sum up, an imbalanced sample frequency causes an imbalance in the norm of each weight vector, and it indicates that there is a discrepancy in the volume of feature space allocated to each class.

The volume discrepancy is in accord with the empirical distribution.
Since a small number of samples are provided from the $K$-th class, the network is trained to allocate a small volume.
Conceptually, this phenomenon is against our desire, since it implies that the network considers more frequent classes as more important classes.
We want to train the network to treat all classes as equally important.
To this end, we propose Weight Vector Normalization (WVN), which normalizes the weight vectors at the end of each training iteration.
Then, the stochastic gradient descent optimizer becomes \textit{projective} stochastic gradient descent optimizer.
From the perspective of the prior distribution, WVN is used to force the class conditional distribution to have the same variance independent of the sample frequency.

\subsection{Generalization}
\label{sec:generalization}
Another important observation of ours is about generalization and the size of the feature cluster.
Higher sample frequency implies a bigger cluster in the image space. 
Even if we consider the effective number of samples~\cite{cui2019class}, the size of the cluster monotonically increases with the number of provided samples.
On the contrary, we have observed that the size of the cluster is \textit{not} monotonic in the feature space, since the feature extraction network is \textit{trained} to project all the samples from each class to a corresponding point.
Moreover, owing to the gap of generalization for each class, the size of the feature cluster monotonically \textit{decreases} in the test time with the number of samples.

Consider a neural network trained to minimize empirical loss.
There is a unanimous agreement that more training data implies better generalization.
Intuitively, more training data represents a high sampling rate, which is associated with less uncertainty\cite{khan2019striking}.
The same analysis is applicable to each class.
If the empirical distribution of the training dataset is imbalanced, the network would provide better generalization for more frequent classes.
It is intuitively straight-forward since the network had seen more diverse data points from classes with a high sample frequency.
Consequently, for frequent classes, the features of the training and test time form clusters close to each other.
On the contrary, if only a few samples are provided, the over-fitting problem arises.
A feature extraction network projects training samples to the feature space close to each other, while projecting the test samples far apart from the training samples.
The most representative method for resolving the over-fitting problem is to reduce the model capacity.
Unfortunately, it is not practicable for class imbalanced scenario, since the reduction of model capacity will deteriorate the performance of other frequent classes.
As a result, the network is trained to provide poor generalization for less frequent classes.

\begin{figure}[t]
  \centering
  \includegraphics[width=\linewidth]{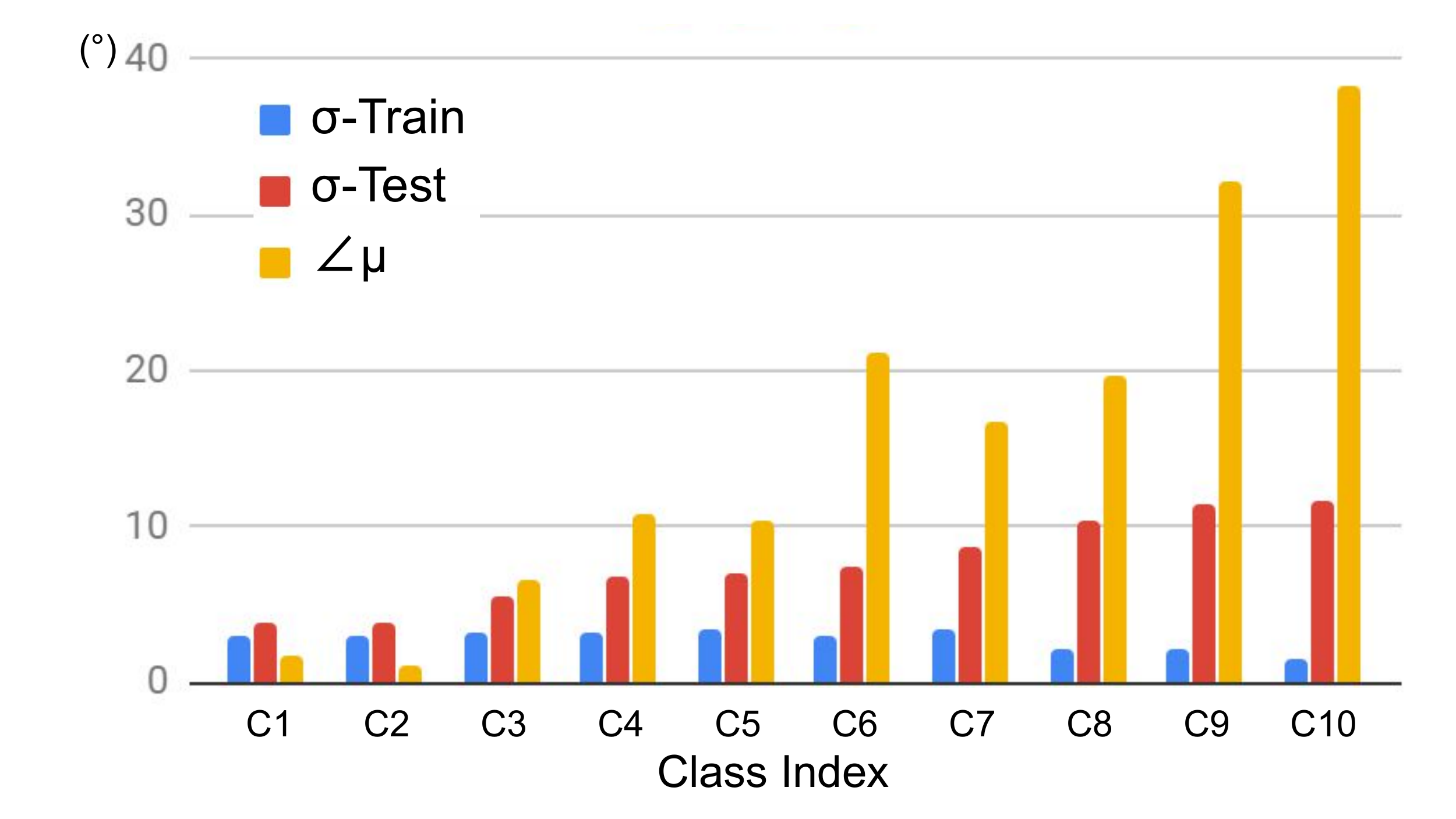}
\vspace{-3 mm}
  \caption{The disparity in generalization depending on the sample frequency.  
  $\sigma$-Train and $\sigma$-Test denote the size of the feature cluster for each class during the training and test times, respectively.
  Although the training features are well-clustered, $\sigma$-test suggests that the test features from less frequent classes are more broadly distributed.
  $\angle \mu$ denotes the angular gap between the centers of the training and test clusters.
  This suggests that the decision boundary should be leaned toward more frequent classes
  }
  \label{fig:generalization}
\vspace{-2 mm}
\end{figure}

Figure~\ref{fig:generalization} presents the cluster size and generalization for each class.
$\sigma$-Train and $\sigma$-Test denote the size of the feature cluster for each class during the training and test times, respectively.
To measure the size of each cluster, we project all the features to unit ball of the feature space.
Then, we calculate the angular standard deviation of each cluster.
Although it is not precisely monotonic, $\sigma$-Train increases with the sample frequency.
Moreover, the size of each cluster becomes saturated if the class has sufficient training samples.
It concurs with the concept of effective number proposed in \cite{cui2019class}.
However, when it comes to the test time, the size of the cluster shows the opposite tendency;
$\sigma$-test suggests that the features from less frequent classes are more broadly distributed, forming larger clusters with lower density.
This shows that the network is well generalized for frequent classes, whereas it is over-fitted for infrequent classes.

The disparity in generalization is more evident when we measure the training/test difference of clusters.
In Figure~\ref{fig:generalization}, $\angle \mu$ of each class denotes the angular gap between the centers of training and test clusters.
We consider this as a measure of generalization.
It shows how far apart the cluster centers are placed during training and test times.
The gap represents the distinctive correlation with the sample frequency.
In the case of the least frequent class, C10, the training and test clusters are nearly 40$^{\circ}$ apart.
Since the last layer of the feature extraction networks is ReLU activation, the maximum angular distance between two feature vectors is 90$^\circ$.
Considering this, we can roughly conceive the significance of the angular gap between training and test clusters.

\begin{figure}[t]
  \centering
  \includegraphics[width=\linewidth]{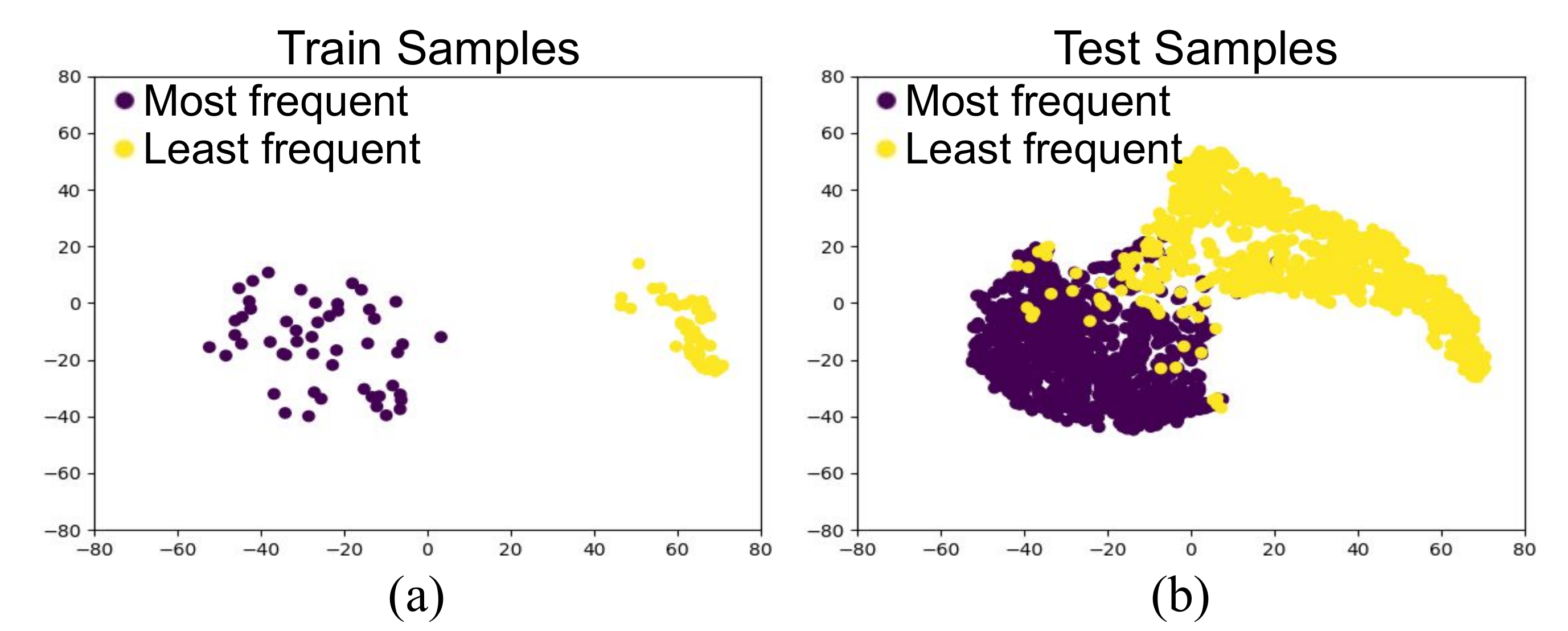}
  \caption{T-sne plot of features from the most and the least frequent classes during the (a) training and (b) test times.
  Same number of samples are plotted from both classes.
  The distribution of the most frequent class is identical for the training and test times, whereas it is distributed far apart for the least frequent class
  }
  \label{fig:tsne}
\end{figure}

Figure~\ref{fig:tsne} visually describes the disparity of generalization.
It is a t-sne plot of the most and the least frequent classes from Long-Tailed CIFAR-10 with an imbalance ratio of 100.
For brevity, the same number of samples are plotted from both classes.
We can visually verify the disparity in generalization.
For the most frequent class, the test features are distributed similar to the training features, while the test features from the least frequent class are more broadly distributed covering the training feature distribution.
Naturally, the center of the test cluster is far apart from the center of the training cluster.

\begin{table*}[t]
\centering
\begin{tabular}{l|c|c|c|c|c|c||c|c|c|c|c|c}
\hline
Dataset       &\multicolumn{6}{c||}{Long Tailed CIFAR-10}     & \multicolumn{6}{c}{Long Tailed CIFAR-100} \\ \hline
Imbalance     & 200   & 100   & 50    & 20    & 10    &  1    & 200   & 100   & 50    & 20    &  10   & 1  \\ \hline
Baseline      & 35.67 & 29.71 & 22.91 & 16.04 & 13.26 &  6.83 & 64.21 & 60.38 & 55.09 & 48.93 & 43.52 & 29.69 \\
Over-sample   & 32.19 & 28.27 & 21.40 & 15.23 & 12.24 &  6.61 & 66.39 & 61.53 & 56.65 & 49.03 & 43.38 & 29.41 \\
Under-sample  & 66.91 & 60.06 & 48.45 & 21.34 & 15.04 &  6.83 & 95.78 & 93.86 & 89.15 & 77.65 & 61.88 & 29.69 \\
Re-weighting  & 38.00 & 31.48 & 23.84 & 17.64 & 13.11 &  6.83 & 75.49 & 71.50 & 59.43 & 51.36 & 43.95 & 29.69 \\
Focal loss \cite{lin2017focal}    & 34.71 & 29.62 & 23.28 & 16.77 & 13.19 &  6.60 & 64.38 & 61.31 & 55.68 & 48.05 & 44.22 & \underline{28.52} \\
CB \cite{cui2019class}  & 31.11 & 25.43 & 20.73 & 15.64 & 12.51 &  6.36 & 63.77 & 60.40 & 54.68 & 47.41 & 42.01 & \textbf{28.39} \\
LDAM  \cite{cao2019learning}& 28.09 & 22.97 & 17.83 & 14.53 & \underline{11.84} &  9.13 & 61.73 & 57.96 & 52.54 & 47.14 & \underline{41.29} & 28.85 \\ \hline
Baseline+RS   & \textbf{27.02} & \underline{21.36} & \underline{17.16} & \underline{13.46} & 11.86 &  \underline{6.32} & \underline{59.59} & \underline{55.65} & \underline{51.91} & \textbf{45.09} & 41.45 & 29.80\\
WVN+RS        & \underline{27.23} & \textbf{20.17} & \textbf{16.80} & \textbf{12.76} & \textbf{10.71} &  \textbf{6.29} & \textbf{59.48} & \textbf{55.50} & \textbf{51.80} & \underline{46.12} & \textbf{41.02} & 29.22 
\\ \hline
\end{tabular}
\caption{Validation errors of ResNet-32 on Long-Tailed CIFAR datasets with various imbalance ratios.
The best performance is denoted in bold, and the second-best performance is underlined.
The results show that our proposed method outperforms the existing methods.
Interestingly, we can achieve better performance than the existing methods by simply re-scaling the weight vectors of the baseline model
}
\label{table:cifar_performance}
\end{table*}

These observations suggest that the decision boundary should rather be leaned toward classes with a high sample frequency, thereby allocating a smaller volume.
This is the opposite tendency of what the minimization of empirical loss induces.
A similar analysis appears in \cite{cao2019learning,khan2019striking}, where the authors suggest that we should encourage a bigger margin for minority classes and propose a novel loss function based on their suggestions.
Since Figure~\ref{fig:norm_and_boundary} shows that we can adjust the decision boundary by controlling the norm of the weight vectors, we propose to re-scale the weight vectors as follows:
\begin{equation}
\label{eq:weight_scaling}
    w_i \leftarrow (\frac{n_{1}}{n_i})^{\gamma}w_{i},
\end{equation}
where $\gamma$ is a hyper-parameter. 
To sum up, our overall training algorithm can be written as follows:
\begin{algorithm}
\caption{Proposed Algorithm}
\label{algo}
\DontPrintSemicolon
Require: dataset $\mathcal{D} = \{x_i\}_{x=1}^{N}, \gamma, \eta$ \;
Initialize $f(\cdot;\theta), W$ \;
\While{training}{
Sample mini-batch $\mathcal{M}$ from $\mathcal{D}$\;
Compute gradient and update: \;
$\theta \leftarrow \theta - \eta\nabla_{\theta} \mathcal{L}(\mathcal{M})$\;
$W \leftarrow W - \eta\nabla_{W} \mathcal{L}(\mathcal{M})$\;
Normalize weight vectors: $\forall i, w_{i} \leftarrow  \frac{w_{i}}{||w_{i}||}$\;
}
Re-scale weight vectors: $\forall i, w_{i} \leftarrow  (\frac{n_{1}}{n_i})^{\gamma}w_{i}$
\end{algorithm}

\newpage
\noindent Note that if $\gamma=0$, all the weight vectors remain the same, ablating the re-scaling method.
With a larger value of $\gamma$, we allocate more volume of feature space to infrequent classes, admitting that our network is poorly generalized for those classes.

\section{Experiments}
In this section, we present the experimental results and analysis.
We evaluate our proposed methods on the object classification task with modified CIFAR~\cite{krizhevsky2009learning} and Tiny ImageNet~\cite{russakovsky2015imagenet} datasets. 
The proposed weight vector normalization is denoted as WVN, and the re-scaling method is denoted as RS.

\subsection{Visual recognition on CIFAR}
\label{sec:result_cifar}
The CIFAR dataset originally contains 50,000 training images and 10,000 test images.
Since the dataset provides a well-balanced empirical distribution, we need to artificially implant the imbalance.
To verify our algorithm and compare with the result of previous research, we applied the long tailed imbalance implanting protocol proposed in \cite{cui2019class}.
The number of training samples decreases according to an exponential function, while the whole test samples were used as it is.
This suggests that a network should be trained to recognize every class regardless of their sample frequency.
Moreover, we used the imbalanced CIFAR dataset for further analysis;
the characteristic of the decreasing number of samples allows us to analyze whether a tendency is dependent on the sample frequency or not.
We used this dataset for the figures in prior sections as well.
For the network architecture, we used ResNet-32~\cite{DBLP:journals/corr/HeZRS15} for all the experiments on CIFAR.
We trained the network over 180 epochs with an initial learning rate of 0.1. 
The learning rate was decayed by a factor of 0.1 at the 80$^{th}$ and 150$^{th}$ epochs.

\begin{figure}[t]
  \centering
  \includegraphics[width=\linewidth]{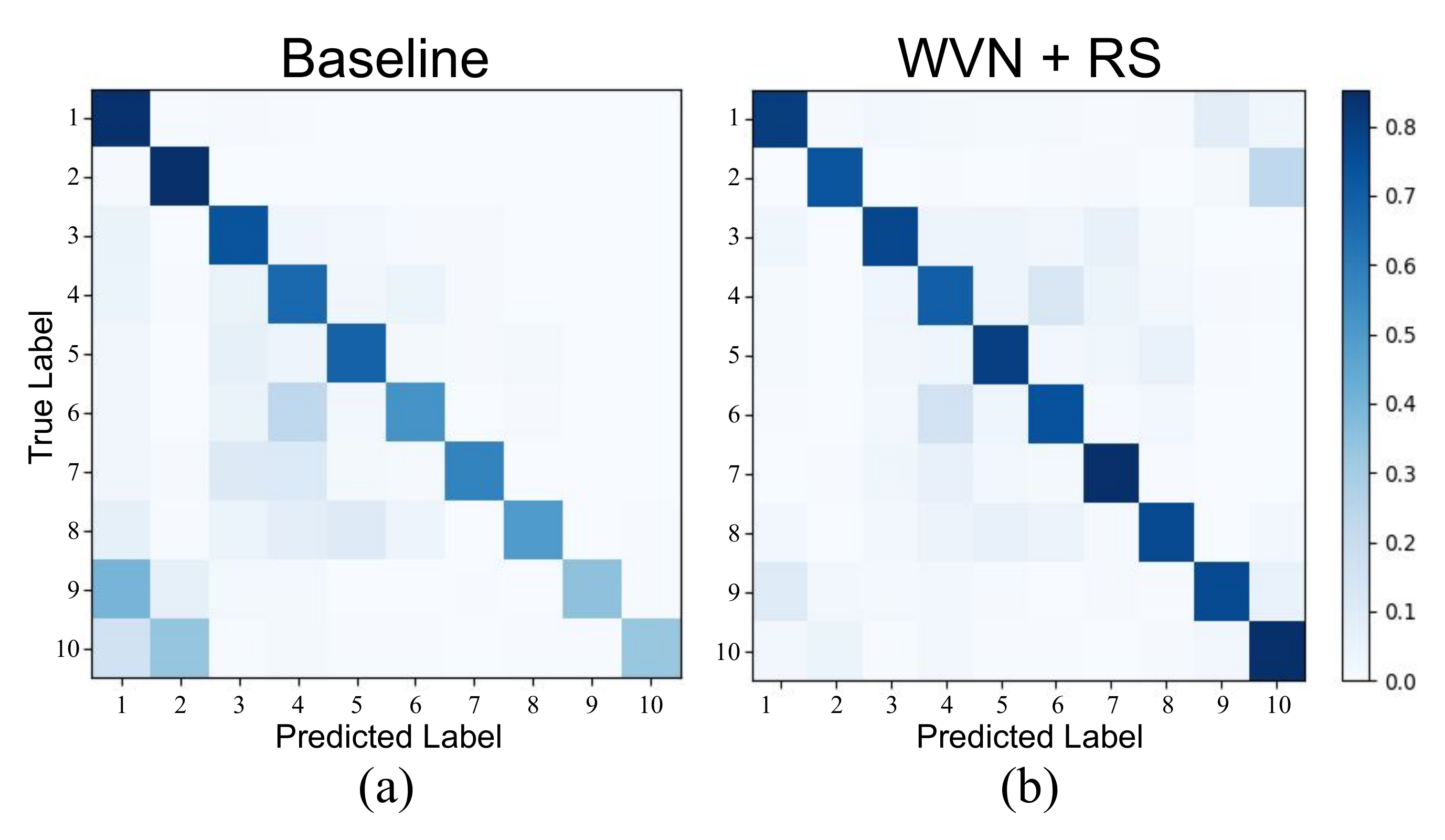}
  \caption{Confusion matrix of the (a)baseline and (b)proposed method on Long Tailed CIFAR-10 with an imbalance ratio of 100.
  The fading color of diagonal elements in (a) implies the disparity of the accuracy.
  With our proposed algorithm, the performance on infrequent classes is improved while preserving the performance on frequent classes
  }
  \label{fig:confmat}
\end{figure}

Table~\ref{table:cifar_performance} summarizes the classification error rates for the long tailed CIFAR dataset.
As a baseline algorithm, a network is trained by minimizing the empirical cross-entropy loss without any regularization.
The under-sampling strategy severely degrades the performance when the dataset is highly imbalanced.
Moreover, the re-weighting approach was neither effective with high imbalance ratio.
The results show that our proposed method outperforms the other methods when the classes are imbalanced.
If the classes are well balanced, normalizing the norm of each weight vector is the same as not using a multiplicative bias in the classifier network.
It reduces the total degree of freedom and affects the performance.
However, irrespective of whether the performance was improved or degraded, the variation was marginal.

Figure~\ref{fig:confmat} presents the confusion matrices of our baseline and WVN+RS model on Long-Tailed CIFAR-10 with an imbalance ratio of 100.
In Figure~\ref{fig:confmat} (a), the color of the diagonal elements is fading.
This shows that the accuracy increases with the number of samples, suggesting that the different sample frequency induces the disparity of the accuracy.
On the other hand, the high values in the bottom left corner represent the low precision of frequent classes and the low recall of the infrequent classes.
It implies that the decision boundary is leaned toward minority classes, while the feature points are biased toward majority classes.
Figure~\ref{fig:confmat} (b) shows that the WVN+RS method alleviates the disparity.
Compared to the baseline, the model trained using our method provides more balanced accuracy.
The performance on infrequent classes is improved while preserving the performance on frequent classes.

Furthermore, the most striking result is that of Baseline+RS, which is the off-the-shelf proxy of our proposed method.
Algorithm~\ref{algo} shows that each weight vector needs to be normalized at the end of every training iteration.
Instead, we only apply Eq.(\ref{eq:weight_scaling}) after the network is trained.
The parameters of the classifier are re-scaled only once \textit{after} the whole training is done.
In other words, we have ablated the weight vector normalization.
Therefore, all the parameters except those in the classifier have an identical value with that of baseline model; it uses identical features with that of the baseline model.
The direction of each weight vector is also preserved from the baseline as well.
Notably, it shows better performance than the other methods.
This suggests that the features extracted by baseline models are of satisfactory quality.

\subsection{Visual recognition on Tiny ImageNet}
We also evaluated the proposed method with Tiny ImageNet~\cite{russakovsky2015imagenet}.
The Tiny ImageNet dataset has 200 classes, and each class has 500 training samples and 50 test samples.
To implant the data imbalance, we applied the same protocol used for CIFAR dataset.
In addition, a step imbalance~\cite{buda2018systematic} was implanted to verify whether our proposed method can resolve various types of imbalance.
In step imbalance, all the majority classes have the same number of samples.
All the minority classes also have the same number of samples but are fewer.
Half of the classes were selected as the minority classes.
We used ResNet-18 architecture, and $\gamma$ was fixed as 0.1 for all the experiments.

\begin{table}[t]
\centering
\begin{tabular}{l|cc|cc}
\hline
Imbalance Type  &\multicolumn{4}{c}{Long Tailed} \\ \hline
Imbalance Ratio &\multicolumn{2}{c|}{100}&\multicolumn{2}{c}{10}\\ \hline
Method          & Top-1 & Top-5 & Top-1 & Top-5   \\ \hline
Baseline        & 66.19 & 42.63 & 50.33 & 26.68   \\
Over Sample     & 71.15 & 52.86 & 54.01 & 31.55   \\
CB Focal \cite{cui2019class}       & 72.72 & 52.62 & 51.58 & 28.91   \\ 
LDAM    \cite{cao2019learning}        & 62.53 & 39.06 & \underline{47.22} & \underline{23.84}   \\ \hline
Baseline+RS     & \underline{62.14} & \underline{37.87} & 48.07 & 24.78   \\
WVN+RS          & \textbf{59.74} & \textbf{36.61} & \textbf{45.77} & \textbf{22.50}   \\ \hline \hline
Imbalance Type  &\multicolumn{4}{c}{Step} \\ \hline
Imbalance Ratio &\multicolumn{2}{c|}{100}&\multicolumn{2}{c}{10}\\ \hline
Method          & Top-1 & Top-5 & Top-1 & Top-5   \\ \hline
Baseline        & 63.82 & 44.09 & 50.89 & 27.06   \\
Over Sample     & 66.78 & 54.82 & 57.23 & 37.35   \\
CB Focal \cite{cui2019class}       & 74.90 & 59.14 & 54.51 & 33.23   \\ 
LDAM    \cite{cao2019learning}        & \underline{60.63} & \underline{38.12} & 47.43 & 23.26   \\ \hline
Baseline+RS     & \textbf{60.07} & \textbf{35.64} & \underline{46.77} & \underline{23.14}   \\
WVN+RS          & 61.26 & 40.36 & \textbf{45.52} & \textbf{22.93}   \\ \hline
\end{tabular}
\caption{Validation errors of ResNet-18 on Tiny ImageNet datasets.
The proposed method shows notable improvements.
Baseline+RS model shows remarkable performance on Tiny ImageNet as well
}
\label{table:tiny_performance}
\end{table}

Table~\ref{table:tiny_performance} presents a summary of the validation errors on the Tiny ImageNet dataset.
Similar to that in the results of the imbalanced CIFAR dataset, WVN+RS method showed the best performance in every experiment except for the case of step imbalance with a ratio of 100.
Even in that case, Baseline+RS model performed the best.
The Baseline+RS models showed remarkable performance on the Tiny ImageNet as well.
Considering that the extracted features are completely identical with those of the baseline model, the superior results of Baseline+RS shows the importance of the decision boundary. 
To this end, we further analyze the off-the-shelf proxy of our proposed method.

\begin{table}[t]
\begin{tabular}{l|cc|cc}
\hline
Imbalance Ratio    & \multicolumn{2}{c|}{100}                      & \multicolumn{2}{c}{10}                         \\ \hline
Feature Extractor  & \multicolumn{1}{c}{Baseline} & \multicolumn{1}{c|}{WVN} & \multicolumn{1}{c}{Baseline} & \multicolumn{1}{c}{WVN} \\ \hline
+RS                & 21.36                   & 20.17                   & 11.86                   & 10.71                  \\ 
+Oracle            & 17.75                   & 18.65                   &  9.94                   & 10.92                  \\ \hline
                   & -3.61                   & \textbf{-1.52}          & -1.92                   & \textbf{-0.21}         \\ \hline
\end{tabular}
\caption{Evaluation error of the Oracle and proposed method.
The last row denotes the performance gap between the proposed method and Oracle.
Although the Oracle performance with the baseline feature extractor is superior to that of the WVN model, the results suggests that the feature extractor of the WVN model can achieve better performance when we apply the RS method}
\label{table:upper_bound}
\end{table}

\begin{figure*}[t]
  \centering
  \includegraphics[width=\linewidth]{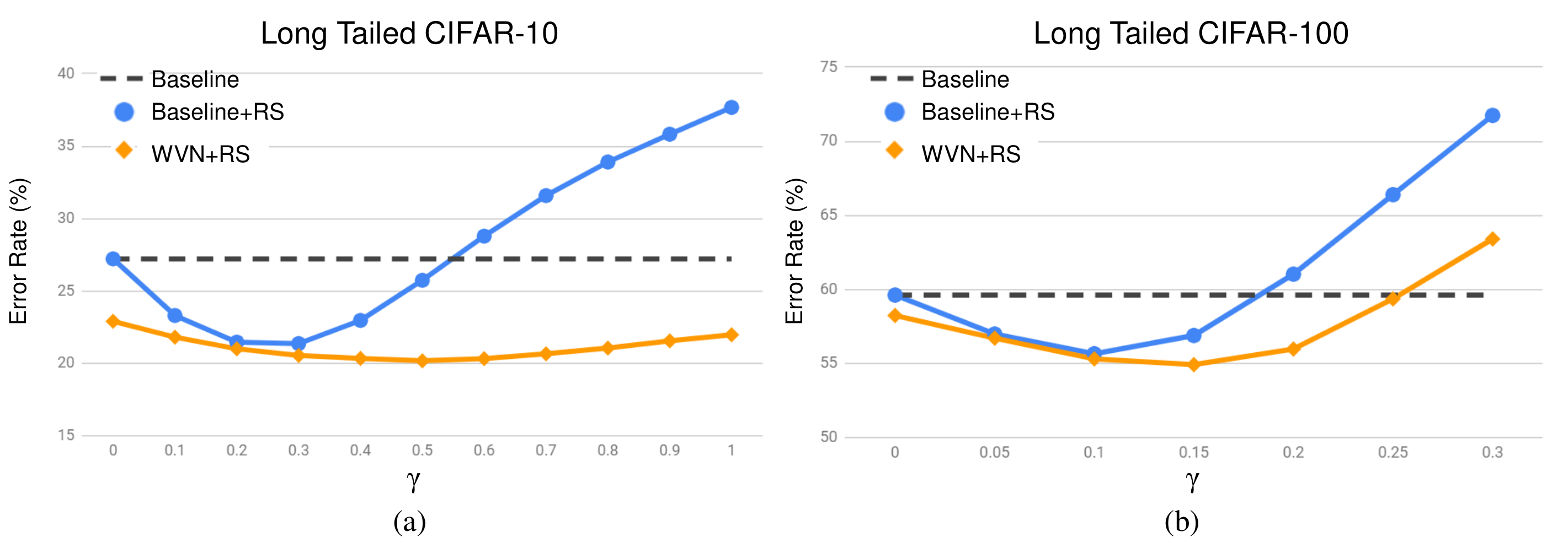}
  \caption{Validation errors of the proposed methods depending on $\gamma$. 
  This figure illustrates that the WVN model is more robust to $\gamma$ than the baseline model
  }
  \label{fig:gamma_sensitivity}
\vspace{-3 mm}
\end{figure*}

\subsection{Discussion}
\label{sec:analysis}
The overall experimental results indicate that
(1) adjustment on the norm of each weight vector can effectively regulate the network to learn from imbalanced data, and (2) the features from the baseline network are already of fine quality.
In particular, by observing the results of Baseline+RS model, we conclude that drawing an appropriate decision boundary is as important as extracting features of superior quality.
The results shown in the previous sections imply that the softmax cross entropy loss is advantageous in training the feature extractor, whereas the resulting classifier provides a biased decision boundary.
To quantify the quality of the extracted features, we have fine-tuned the classifier with test samples, while the feature extraction network is fixed.
We denote these performances as Oracle.
Table~\ref{table:upper_bound} summarizes the validation error of the proposed methods and their Oracle performance depending on the feature extractor.
Since the classifier of the Oracle model is trained and validated with the same test data, their validation error can be interpreted as a lower bound of the corresponding feature extractor.
In Table~\ref{table:upper_bound}, RS denotes the performance after the weight vectors are re-scaled.
Interestingly, the Oracle of the Baseline feature extractor performs better than the Oracle of WVN in both cases.
This suggests that the feature extractor is rather degraded by the weight vector normalization in terms of the potential performance.
Nevertheless, the WVN+RS model performs better than the Baseline+RS model.
The last row of Table~\ref{table:upper_bound} shows the performance gap between the proposed method and their Oracle.
Note that the improved performance by adding the RS method can approach much closer to the Oracle performance if the vector is normalized.
This shows that the features from the WVN model are aligned more appropriately, so that we can draw better decision boundary.

More important benefit of WVN is $\gamma$ sensitivity.
Figure~\ref{fig:gamma_sensitivity} shows that the Baseline+RS model is more sensitive in the selection of $\gamma$.
Note that if $\gamma$ is zero, Baseline+RS is the same as Baseline.
A larger $\gamma$ denotes a stronger adjustment on the decision boundary.
Therefore, the robustness with regard to $\gamma$ implies that the features are clustered with a large margin in the test time.
The robustness is also important in terms of selecting the hyper-parameter.
Since the proposed weight re-scaling is a post-processing of the training procedure, adjusting $\gamma$ is relatively handy compared with other cost-sensitive methods.
Nevertheless, it is clearly advantageous that we can determine the hyper-parameter effortlessly.

From another viewpoint, the models are more sensitive on Long Tailed CIFAR-100 than CIFAR-10.
In the experiments on Long Tailed CIFAR-10, the proposed method consistently showed better performance than the baseline.
Moreover, the variation of performance is marginal along the value of $\gamma$.
However, on CIFAR-100, the robustness against $\gamma$ was comparatively degraded.
A small interval of $\gamma$ improves the performance, and a larger value degrades the performance.
This is understandable since the classes in CIFAR-100 are more fine-grained than the classes in CIFAR-10.
Fine-grained classes are highly likely to have a smaller margin than coarse-grained classes.
Since the proposed re-scaling method is effective when the decision boundary is adjusted in between the margin, a small margin causes high $\gamma$-sensitivity.

\section{Conclusion}
In this paper, we proposed two methods for class imbalanced learning: weight vector normalization (WVN) and weight re-scaling (RS).
Our methods showed outstanding performance despite their simplicity.
The key idea of our methods is a causal relationship between the imbalanced sample frequency, norm of the weight vector, and decision boundary.
We showed that the disparity in the norm is a consequence of imbalanced class and described how the disparity affects the decision boundary.
Moreover, we experimentally showed that we could successfully adjust the decision boundary.

Most of the deep learning based methods for class imbalanced learning follow a cost-sensitive approach, seeking a better loss function.
The underlying concept of cost-sensitive methods is to train a better feature extractor.
Although the models trained with those methods perform better than the baseline, this work showed that a simple adjustment on the baseline further improves the performance.
In particular, the results of Baseline+RS model show that the baseline features were already of fine quality.
This suggests that drawing a better decision boundary is as important as training a better feature extractor.
We hope that this work provides a novel viewpoint and inspiration for class imbalanced learning.


\newpage
{\small
\bibliographystyle{ieee_fullname}
\bibliography{egbib}
}


\newpage
\onecolumn
\vspace*{5 mm}
\begin{Large}
\begin{center}
\textbf{Adjusting Decision Boundary for Class Imbalanced Learning \newline
- Supplementary Material}
\end{center}
\end{Large}
\section*{Relative norm and step imbalance}
Figure 2 in our main paper shows the evident correlation between the norm and the sample frequency.
To verify that the tendency is not restricted to the CIFAR dataset with long-tailed imbalance, we present the relative norm of the network trained on TinyImageNet with step imbalance.
Since there are 200 classes, we randomly selected 20 classes from frequent and infrequent classes, respectively.
Note that every frequent classes have 500 training samples.
Infrequent classes of Imb10 and Imb100 have 50 and 5 training samples, respectively.
Similar to the Figure 2 in our main paper, classes with the same sample frequency show fluctuation in the norm.
However, the amplitude is marginal compare to the discrepancy between the frequent and infrequent classes.
This result suggests that the tendency of the norm is clearly induced by the sample frequency.
\begin{figure}[h]
  \centering
  \includegraphics[width=0.5\linewidth]{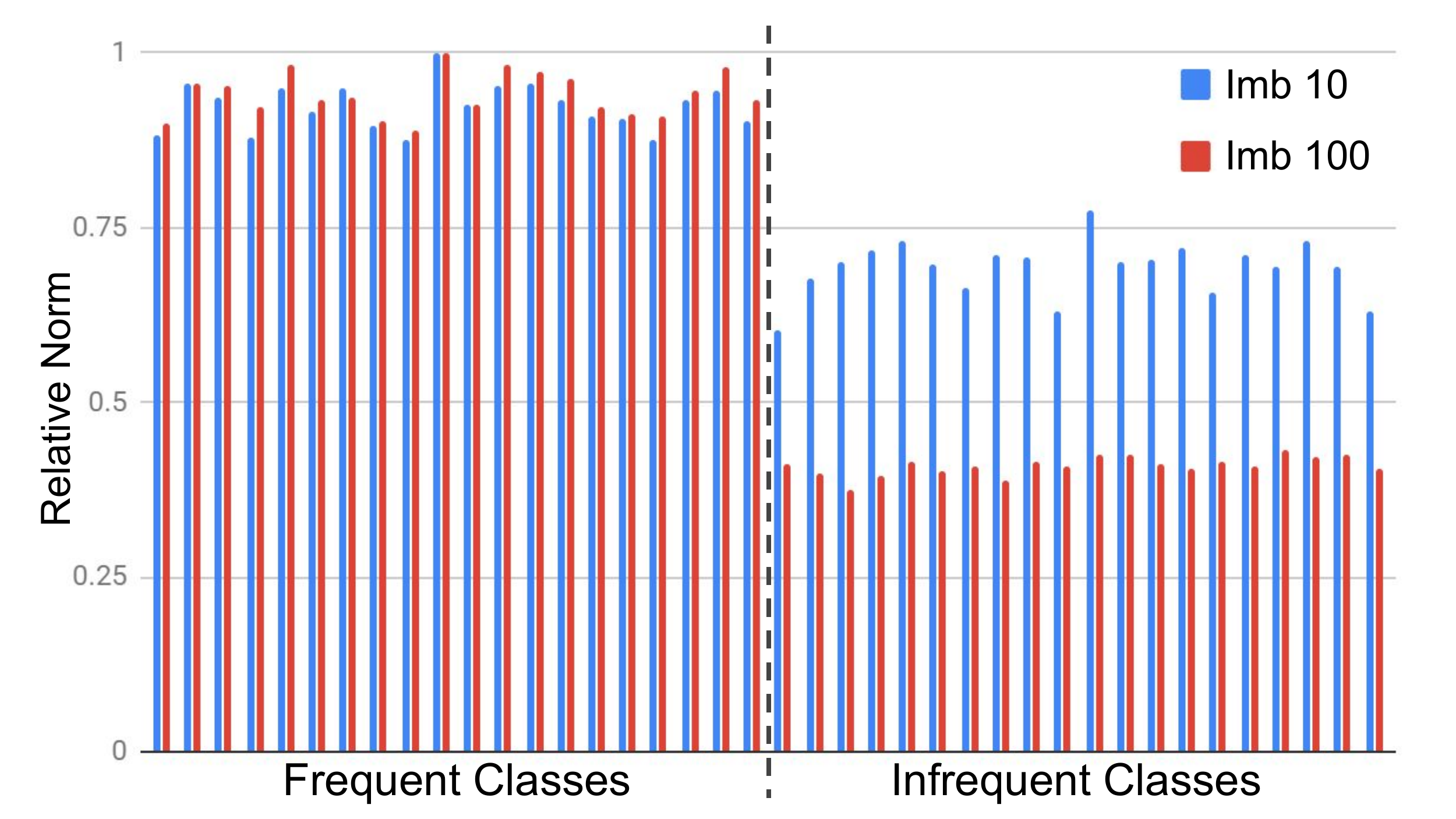}
  \caption{Relative norm of the weight vectors of the network trained on TinyImageNet dataset with step imbalance}
  \label{fig:relative_norm_step}
\end{figure}

\section*{Relative norm and over-sampling}
\begin{figure}[h]
  \centering
  \includegraphics[width=0.5\linewidth]{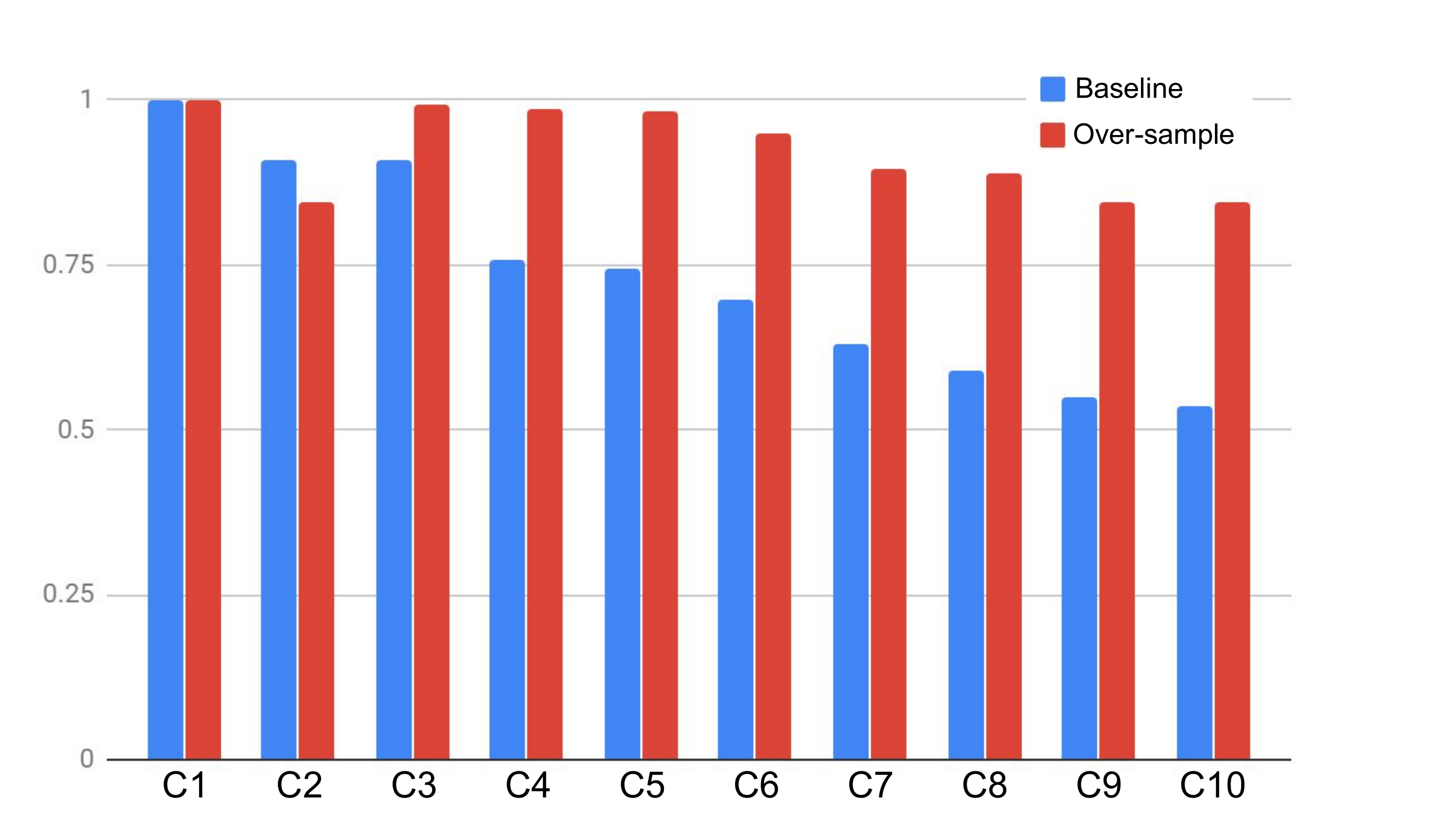}
  \caption{Relative norm of the weight vectors of the network trained with over-sampling strategy.
            The long-tailed CIFAR-10 with imbalance factor of 100 is used for visualization}
  \label{fig:relative_norm_repeat}
\end{figure}

The simplest method balancing the sample frequency is the over-sampling strategy.
Figure~\ref{fig:relative_norm_repeat} shows that the norm discrepancy is remarkably decreased by over-sampling strategy.
Although it is marginal, the results in our main paper (Table 1) suggests that the class imbalance problem is resolved in some cases.
Applying over-sampling strategy is the same as balancing the weights in Eq.(3) in our main paper.
Nevertheless, the norm is still increasing with the sample frequency.
It suggests that the decision boundary is still biased toward less frequent classes.
\vspace{-3 mm}

\end{document}